\documentclass[twoside,11pt]{article}
\pdfoutput=1

\usepackage{jmlr2e}
\usepackage{doi}

\usepackage{subcaption}
\usepackage{url}
\usepackage{color}
\usepackage{listings}
\usepackage{amsmath}

\jmlrheading{?}{2015}{?-?}{?/??}{?/??}{Jaakko Luttinen}

\ShortHeadings{BayesPy: Variational Bayesian Inference in Python}{Luttinen}
\firstpageno{1}

\begin{document}

\title{BayesPy: Variational Bayesian Inference in Python}

\author{\name Jaakko Luttinen \email jaakko.luttinen@aalto.fi \\
       \addr Department of Computer Science\\
       Aalto University, Finland}

\editor{?}

\maketitle

\begin{abstract}%
  BayesPy is an open-source Python software package for performing variational
  Bayesian inference.  It is based on the variational message passing framework
  and supports conjugate exponential family models.  By removing the tedious
  task of implementing the variational Bayesian update equations, the user can
  construct models faster and in a less error-prone way.  Simple syntax,
  flexible model construction and efficient inference make BayesPy suitable for
  both average and expert Bayesian users.  It also supports some advanced
  methods such as stochastic and collapsed variational inference.
\end{abstract}


\section{Introduction}

Bayesian framework provides a theoretically solid and consistent way to
construct models and perform inference.  In practice, however, the inference is
usually analytically intractable and is therefore based on approximation methods
such as variational Bayes (VB), expectation propagation (EP) and Markov chain
Monte Carlo (MCMC) \citep{Bishop:2006}.  Deriving and implementing the formulas
for an approximation method is often straightforward but tedious, time consuming
and error prone.

BayesPy is a Python 3 package providing tools for constructing conjugate
exponential family models and performing VB inference easily and efficiently.
It is based on the variational message passing (VMP) framework which defines a
simple message passing protocol \citep{Winn:2005}.  This enables implementation
of small general nodes that can be used as building blocks for large complex
models.  BayesPy offers a comprehensive collection of built-in nodes that can be
used to build a wide range of models and a simple interface for implementing
custom nodes.  The package is released under the MIT license.

Several other projects have similar goals for making Bayesian inference easier
and faster to apply.  VB inference is available in Bayes Blocks
\citep{Raiko:2007}, VIBES \citep{Bishop:2002} and Infer.NET \citep{Infer.NET}.
Bayes Blocks is an open-source C++/Python package but limited to scalar Gaussian
nodes and a few deterministic functions, thus making it very limited.  VIBES is
an old software package for Java, released under the revised BSD license, but it
is no longer actively developed.  VIBES has been replaced by Infer.NET, which
is partly closed source and licensed for non-commercial use only.  Instead of VB
inference, mainly MCMC is supported by other projects such as PyMC \citep{PyMC},
OpenBUGS \citep{OpenBUGS}, Dimple \citep{Dimple} and Stan \citep{Stan}.  Thus,
there is a need for an open-source and maintained VB software package.

\section{Features}

BayesPy can be used to construct conjugate exponential family models.  The
documentation provides detailed examples of how to construct a variety of
models, including principal component analysis models, linear state-space
models, mixture models and hidden Markov models.  BayesPy has also been used in
two publications about parameter expansion and time-varying dynamics for linear
state-space models \citep{Luttinen:2013,Luttinen:2014}.

Using BayesPy for Bayesian inference consists of four main steps: constructing
the model, providing data, finding the posterior approximation and examining the
results.  The user constructs the model from small modular blocks called nodes.
Roughly, each node corresponds to a latent variable, a set of observations or a
deterministic function.
The inference engine is used to run the message passing algorithm in order to
obtain the posterior approximation.  The resulting posterior can be examined,
for instance, by using a few built-in plotting functions or printing the
parameters of the posterior distributions.

Nodes are the primary building blocks for models in BayesPy.  There are two
types of nodes: stochastic and deterministic.  Stochastic nodes correspond to
probability distributions and deterministic nodes correspond to functions.
Built-in stochastic nodes include all common exponential family distributions
(e.g., Gaussian, gamma and Dirichlet distributions), a general mixture
distribution and a few complex nodes for dynamic variables (e.g., discrete and
Gaussian Markov chains).  Built-in deterministic nodes include a gating node and
a general sum-product node.
In case a model cannot be constructed using the built-in nodes, the
documentation provides instructions for implementing new nodes.

BayesPy is designed to be simple enough for non-expert users but flexible and
efficient enough for expert users.  One goal is to keep the syntax easy and
intuitive to read and write by making it similar to the mathematical formulation
of the model.  Missing values are easy to handle and the variables can be
monitored by plotting the posterior distributions during the learning.  BayesPy
has also preliminary support for some advanced VB methods such as stochastic
variational inference \citep{Hoffman:2013}, deterministic annealing
\citep{Katahira:2008}, collapsed inference \citep{Hensman:2012}, Riemannian
conjugate gradient learning \citep{Honkela:2010}, parameter expansions
\citep{Qi:2007} and pattern searches \citep{Honkela:2003}.  For developers, the
unit testing framework helps in finding bugs, making changes and implementing
new features in a robust manner.

BayesPy can be installed similarly to other Python packages.  It requires Python
3 and a few popular packages: NumPy, SciPy, matplotlib and h5py.  The latest
release can be installed from Python Package Index (PyPI) and detailed
instructions can be found from the comprehensive online
documentation\footnote{\url{http://bayespy.org}}.  The latest development
version is available at
GitHub\footnote{\url{https://github.com/bayespy/bayespy}}, which is also the
platform used for reporting bugs and making pull requests.

\section{Example}

This section demonstrates the key steps in using BayesPy.  An artificial
Gaussian mixture dataset is created by drawing 500 samples from two
2-dimensional Gaussian distributions.  200 samples have mean $[2, 2]$ and 300
samples have mean $[0, 0]$:
\begin{lstlisting}
import numpy as np
N = 500; D = 2
data = np.random.randn(N, D)
data[:200,:] += 2*np.ones(D)
\end{lstlisting}
We construct a mixture model for the data and assume that the parameters, the
cluster assignments and the true number of clusters are unknown.  The model uses
a maximum number of five clusters but the effective number of clusters will be
determined automatically:
\begin{lstlisting}
K = 5
\end{lstlisting}
The unknown cluster means and precision matrices are given Gaussian and Wishart
prior distributions:
\begin{lstlisting}
from bayespy import nodes
mu = nodes.Gaussian(np.zeros(D), 0.01*np.identity(D), plates=(K,))
Lambda = nodes.Wishart(D, D*np.identity(D), plates=(K,))
\end{lstlisting}
The \texttt{plates} keyword argument is used to define repetitions similarly to
the plate notation in graphical models.  The cluster assignments are categorical
variables, and the cluster probabilities are given a Dirichlet prior
distribution:
\begin{lstlisting}
alpha = nodes.Dirichlet(0.01*np.ones(K))
z = nodes.Categorical(alpha, plates=(N,))
\end{lstlisting}
The observations are from a Gaussian mixture distribution:
\begin{lstlisting}
y = nodes.Mixture(z, nodes.Gaussian, mu, Lambda)
\end{lstlisting}
The second argument for the \texttt{Mixture} node defines the type of the
mixture distribution, in this case Gaussian.  The variable is marked as observed
by providing the data:
\begin{lstlisting}
y.observe(data)
\end{lstlisting}
Next, we want to find the posterior approximation for our latent variables.  We
create the variational Bayesian inference engine:
\begin{lstlisting}
from bayespy.inference import VB
Q = VB(y, mu, z, Lambda, alpha)
\end{lstlisting}
Before running the VMP algorithm, the symmetry in the model is broken by
a random initialization of the cluster assignments:
\begin{lstlisting}
z.initialize_from_random()
\end{lstlisting}
Without the random initialization, the clusters would not be separated.  The VMP
algorithm updates the variables in turns and is run for 200 iterations or until
convergence:
\begin{lstlisting}
Q.update(repeat=200)
\end{lstlisting}
The results can be examined visually by using \texttt{bayespy.plot} module:
\begin{lstlisting}
import bayespy.plot as bpplt
bpplt.gaussian_mixture_2d(y, alpha=alpha)
bpplt.pyplot.show()
\end{lstlisting}
It is also possible to print the parameters of the approximate posterior
distributions:
\begin{lstlisting}
print(alpha)
\end{lstlisting}
The \texttt{bayespy.plot} module contains also other functions for visually
examining the distributions.

\section{Comparison with Infer.NET}

This section provides a brief comparison with Infer.NET because it is another
active project implementing the variational message passing algorithm, whereas
the other related active projects implement MCMC.  The main advantages of
Infer.NET over BayesPy are its support for non-conjugate models (e.g., logistic
regression) and other inference engines (EP and Gibbs sampling).  On the other
hand, BayesPy is open-source software and supports several advanced VB methods
(e.g., stochastic variational inference and collapsed inference).

The speed of the packages were compared by using two widely used models: a
Gaussian mixture model (GMM) and principal component analysis
(PCA).\footnote{The scripts for running the experiments are available as
  supplementary material.}  Both models were run for small and large artificial
datasets.  For GMM, the small model used 10 clusters for 200 observations with 2
dimensions, and the large model used 40 clusters for 2000 observations with 10
dimensions.  For PCA, the small model used 10-dimensional latent space for 500
observations with 20 dimensions, and the large model used 40-dimensional latent
space for 2000 observations with 100 dimensions.  For the PCA model, Infer.NET
used both a fully factorizing approximation and also the same approximation as
in BayesPy which did not factorize with respect to the latent space dimensions.
The experiments were run on a quad-core (i7-4702MQ) Linux computer.
Because the packages run practically identical algorithms and thus
converged to similar solutions in a similar number of iterations (50--100
iterations depending on the dataset), we compared the average CPU time per
iteration.

The results are summarized in Table~\ref{tab:speed}.  For all datasets, BayesPy
is faster probably because it uses highly optimized numerical libraries (BLAS
and LAPACK).  For PCA, BayesPy also automatically avoids redundant computations
which arise because latent variables have equal posterior covariance matrices.
However, such broadcasting is not always possible (e.g., if there are missing
values in the PCA data).  Thus, the table also presents the results when BayesPy
is forced to not use broadcasting: BayesPy is slower than Infer.NET on the
smaller PCA dataset but faster on the larger PCA dataset.
If Infer.NET used the same factorization for PCA as BayesPy, Infer.NET may be
orders of magnitude slower.

\begin{table}[tb]
  \centering
  \caption{
    The average CPU time in milliseconds per iteration for each
    dataset.  The results in parentheses have the following meanings: a) BayesPy without using
    broadcasting. b) Infer.NET using the same factorization as in
    BayesPy.
  }
  \small
  \begin{tabular}{ccccc}
    &
    Small GMM
    &
    Large GMM
    &
    Small PCA
    &
    Large PCA
    \\
    \hline
    BayesPy     & 6 & 90  & 7 (60) & 400 (1\,500)
    \\
    Infer.NET   & 25 & 4\,600 & 37 (350) & 2\,200 (210\,000)
  \end{tabular}
  \label{tab:speed}
\end{table}

\section{Conclusions}

BayesPy provides a simple and efficient way to construct conjugate exponential
family models and to find the variational Bayesian posterior approximation in
Python.
In addition to the standard variational message passing, it supports several
advanced methods such as stochastic and collapsed variational inference.  Future
plans include support for non-conjugate models and non-parametric models (e.g.,
Gaussian and Dirichlet processes).

\vskip 0.2in
\small
\bibliography{bibliography/bibliography.bib}

\end{document}